\documentclass[runningheads,a4paper]{llncs}

\usepackage{amssymb}
\usepackage{graphicx}

\usepackage{url}
\usepackage{booktabs}
\usepackage{multirow}
\usepackage{soul}
\usepackage{amsmath}
\usepackage{marvosym}

\begin{document}

\mainmatter  

\title{MambaHand: Local Identity and Behavioral Sequence Learning with Mamba for Dynamic Hand Gesture Authentication}


%
%
%


%
%

\author{
	Zenan Lin\inst{1} \and
	Junduan Huang\inst{2}\textsuperscript{\Letter} \and
	Xilai Wang\inst{1} \and
	Wenxiong Kang\inst{1}
}

\institute{
	School of Automation Science and Engineering, South China University of Technology, Guangzhou, China
	\and
	School of Artificial Intelligence, South China Normal University, Foshan, China \\
	\email{runrunjun@163.com}
}

\authorrunning{Z. Lin et al.}
\titlerunning{MambaHand: Dynamic Hand Gesture Authentication}

\maketitle

\begin{abstract}
\label{Abstract}

Dynamic hand gesture authentication (DHGA) is an emerging biometric technique that contains both physiological and behavioral identity characteristics. However, existing methods primarily rely on convolutional networks or manually designed motion representations, making it difficult to jointly capture fine-grained local identity differences and long-range behavioral dependencies during hand gesture execution. To address these challenges, we propose a two-stream identity representation learning framework, MambaHand Net, to combine CNNs with Mamba for DHGA. The proposed framework consists of a Local Identity Stream (LIS) and a Behavioral Sequential Stream (BSS). The LIS extracts frame-wise local identity cues with a convolutional network, while the BSS preserves spatial tokens from intermediate visual features and constructs first-order and second-order temporal difference sequences to model multi-order behavioral variations. Moreover, the bidirectional Mamba is incorporated into the BSS to model semantic motion across different spatial locations, enabling efficient long-range modeling of hand gesture execution habits. Finally, attention-guided spatial token aggregation and a two-stream fusion strategy are used to integrate local identity cues and multi-order behavioral patterns into a discriminative identity representation. Our method achieves state-of-the-art (SOTA) average performance on the SCUT-DHGA benchmark dataset, demonstrating that combining CNNs with Mamba is an effective new solution for DHGA.


\end{abstract}

\section{Introduction}
Dynamic hand gesture authentication (DHGA) is an emerging visual biometric technique that verifies users based on hand gesture videos \cite{liu2020dynamic,lin2025multiscale,wu2015leveraging}. Compared with passwords and smart cards, dynamic gestures do not require users to memorize complex secrets or carry physical credentials. Similar to face, finger vein \cite{huang2025study,huang2026study}, palmprint \cite{qiao2025normalized}, and gait biometrics \cite{chao2021gaitset}, DHGA also enables natural, convenient, and contactless authentication. Since dynamic gestures contain both physiological cues, such as hand shape, texture, and local posture, and behavioral cues, such as speed and acceleration, effectively integrating local identity information and long-term behavioral habits is crucial for DHGA \cite{wu2016two,liu2020dynamic,lin2025multiscale}.

Different from ordinary hand gesture recognition \cite{abavisani2019improving}, which focuses on recognizing ``which gesture is performed'', and hand gesture reconstruction \cite{li2023chord,zhan2024oakink2}, which targets recovering hand mesh geometry and joint keypoints regardless of subject identity, DHGA aims to determine ``who performs the gesture'' \cite{abavisani2019improving,song2022video}. Therefore, the model needs to capture fine-grained identity differences among users performing similar gestures. Wu et al. introduced the HandLogin dataset as an early attempt for DHGA \cite{wu2015leveraging}. Liu et al. constructed the large-scale SCUT-DHGA dataset and established systematic evaluation protocols, providing an important benchmark for subsequent studies \cite{liu2020dynamic}.

Recently, state-space models (SSMs), especially Mamba, have shown strong potential for long-sequence modeling due to their selective state-space mechanism and linear complexity in sequence length. Recent visual Mamba models further demonstrate their effectiveness in modeling temporal dependencies in visual signals~\cite{gu2021efficiently,gu2023mamba,zhu2024vision,luo2024physmamba}. For DHGA, speed variation, acceleration pattern, and hand-region coordination naturally form long-range behavioral sequences. This motivates us to explore Mamba as a behavioral sequence learner for modeling gesture execution habits and fine-grained behavioral differences.

Based on these observations, we propose the MambaHand Net, which jointly models local identity cues and long-range behavioral dynamics for dynamic hand gesture authentication. The framework consists of a Local Identity Stream (LIS) and a Behavioral Sequential Stream (BSS), where multi-order temporal differences and bidirectional Mamba are introduced for behavior modeling. The main contributions are summarized as follows:

\begin{itemize}
	\item We design a two-stream CNN-Mamba framework for DHGA and propose a Temporal Difference Mamba Branch, which constructs multi-order behavior sequences and utilizes bidirectional Mamba to perform long-range modeling of behavior sequences at different spatial locations.
	
	\item We propose an Attention-Guided Behavioral Token Aggregation module, which highlights the identity-discriminating local hand dynamic regions through an attention-guided mechanism while preserving global behavior information in the BSS.
	
	\item We conduct extensive experiments on the SCUT-DHGA dataset and achieves SOTA average authentication performance.
\end{itemize}

\section{RELATED WORK}
\subsection{Dynamic Hand Gesture Authentication}

Dynamic hand gesture authentication aims to verify users from hand gesture videos. Early vision-based methods mainly relied on motion cues such as hand trajectories, geometric structures, and optical flow, while recent studies have adopted deep networks for end-to-end identity representation learning \cite{aumi2014airauth,wu2016two}. Wu et al. proposed TS-CNNs to learn user-specific gesture execution styles with a two-stream convolutional network~\cite{wu2016two}. Liu et al. constructed the large-scale SCUT-DHGA dataset and introduced DHGA-net to model gesture videos with 3D convolutions~\cite{liu2020dynamic}. Song et al. designed temporal difference maps and a 3D TDS Net to enhance behavioral modeling and physiological-behavioral feature fusion~\cite{song2022video}. Lin et al. further proposed multiscale dynamic hand gesture super-images to represent multiscale local and global motion cues~\cite{lin2025multiscale}. 

\subsection{State Space Models}

State Space Models (SSMs) ~\cite{gu2021efficiently,gu2023mamba,zhu2024vision} have recently attracted increasing attention for long-sequence modeling. S4 improves long-term dependency modeling through structured state-space parameterization~\cite{gu2021efficiently}, and Mamba introduces a selective state-space mechanism to dynamically select and propagate input-dependent information~\cite{gu2023mamba}. Compared with Transformer self-attention, Mamba has less complexity with respect to sequence length, making it suitable for efficient long-range modeling. Recent Mamba-based visual models have shown promising performance in image classification, video understanding, and semantic segmentation~\cite{gu2023mamba,zhu2024vision,luo2024physmamba}.

\section{Methodology}

\subsection{Preliminaries}
\textbf{State Space Models (SSMs)} provide an efficient sequence modeling paradigm by representing historical information with a latent state. Given a one-dimensional input sequence $x(t) \in \mathbb{R}$, the continuous-time SSM maps it to an output sequence $y(t) \in \mathbb{R}$ through a hidden state $h(t) \in \mathbb{R}^{N}$:
\begin{equation}
	h'(t) = A h(t) + Bx(t), \quad y(t)=Ch(t),
\end{equation}
where $A \in \mathbb{R}^{N \times N}$ is the state transition matrix, and $B \in \mathbb{R}^{N \times 1}$ and $C \in \mathbb{R}^{1 \times N}$ are the projection matrices, respectively. To apply SSMs to discrete sequences, a time-scale parameter $\Delta$ is used to discretize the continuous system. With the zero-order hold (ZOH) rule, the continuous parameters $A$ and $B$ are transformed into their discrete counterparts $\bar{A}$ and $\bar{B}$:
\begin{equation}
	\bar{A}=\exp(\Delta A), \quad
	\bar{B}=(\Delta A)^{-1}(\exp(\Delta A)-I)\cdot \Delta B.
\end{equation}
The discretized SSM can then be computed in a recurrent form:
\begin{equation}
	h_t = \bar{A}h_{t-1} + \bar{B}x_t, \quad y_t = Ch_t.
\end{equation}
It can also be equivalently written as a global convolution:
\begin{equation}
	\bar{K}=(C\bar{B}, C\bar{A}\bar{B}, \ldots, C\bar{A}^{M-1}\bar{B}), \quad
	y=x*\bar{K},
\end{equation}
where $M$ is the length of the input sequence, $\bar{K}\in\mathbb{R}^{M}$ denotes the structured convolution kernel, and $*$ denotes the convolution operation.

Mamba extends conventional time-invariant SSMs with a selective state space mechanism, where $\Delta$, $B$, and $C$ are generated in an input-dependent manner. This allows the model to selectively propagate useful information while maintaining linear complexity with respect to sequence length.

\begin{figure*}
	\centering
	\setlength{\belowcaptionskip}{-0.2cm}
	\includegraphics[scale = 0.3]{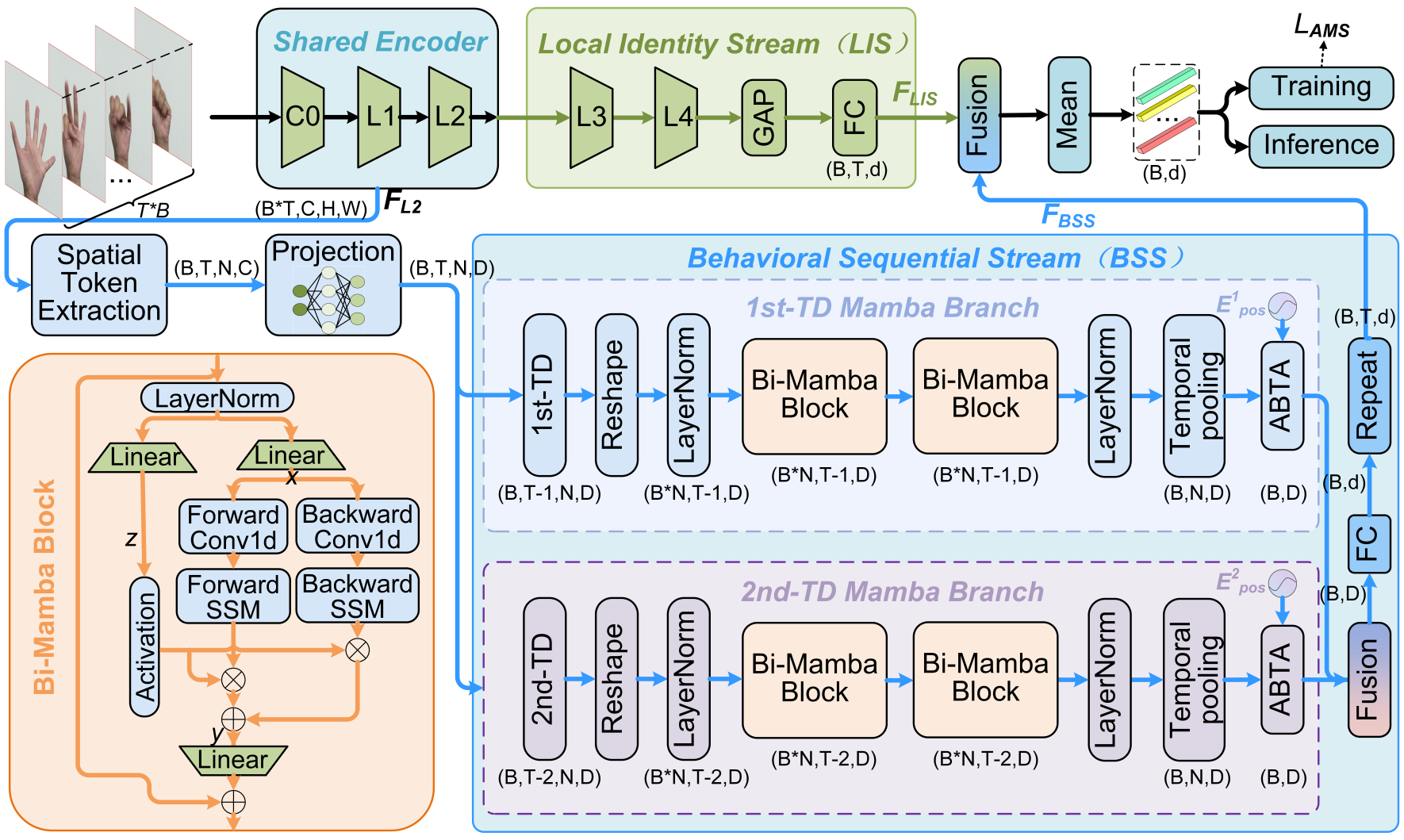}
	\caption{Overall architecture of the proposed MambaHand Net. The framework consists of a LIS and a BSS built upon a shared encoder. The LIS extracts frame-level local identity cues, while the BSS preserves spatial tokens from intermediate features and constructs first-order and second-order temporal difference sequences for behavioral modeling. Two Bi-Mamba-based branches are employed to capture multi-order behavioral dynamics, followed by ABTA, branch fusion, and two-stream feature fusion for final identity representation learning.}
	\label{fig:framework}
	\vspace{-1.0cm}
\end{figure*}

\subsection{Overall Architecture}
Fig.~\ref{fig:framework} shows the overall architecture of the proposed MambaHand Net. Given a batch of $B$ dynamic hand gesture videos, each video contains $T$ sampled frames, denoted as $\mathbf{X}=\{I_1,I_2,\ldots,I_T\}$, where $I_t\in\mathbb{R}^{3\times224\times224}$ is the $t$-th RGB frame. The input frames are first fed into a shared ResNet18 encoder for visual feature extraction. The shared encoder contains an initial convolutional stem $C0$ and two residual stages \textit{layer1} ($L1$) and \textit{layer2} ($L2$).

After $L2$, the intermediate feature map is denoted as $\mathbf{F}_{\mathrm{L2}}\in\mathbb{R}^{BT\times C\times H\times W}$, where $C=128$ and $H=W=28$. This feature is used in two ways. First, it is further processed by $L3$, $L4$, a global average pooling (GAP) layer, and a fully connected layer ($\mathrm{FC}_{\mathrm{LIS}}$) in the Local Identity Stream (LIS) to extract stable identity cues, such as hand shape, texture, and local posture. The output frame-level identity feature is denoted as $\mathbf{F}_{\mathrm{LIS}}\in\mathbb{R}^{B\times T\times d}$, where $d=512$ is the final identity feature dimension.

Second, $\mathbf{F}_{\mathrm{L2}}$ is fed into the Behavioral Sequential Stream (BSS) for behavioral modeling. Specifically, adaptive average pooling is first used to resize the spatial size to $P\times P$. In this work, $P=7$, and thus the pooled feature is $\mathbf{F}_{\mathrm{pool}}\in\mathbb{R}^{BT\times C\times 7\times 7}$. It is then reshaped into a spatial token sequence $\mathbf{Z}\in\mathbb{R}^{B\times T\times N\times C}$, where $N=P^2=49$ denotes the number of spatial tokens. Each token corresponds to a local region of the hand feature map. A linear projection is further used to map the token dimension from $C$ to the Mamba hidden dimension $D=256$, obtaining $\mathbf{Z}_{\mathrm{proj}}\in\mathbb{R}^{B\times T\times N\times D}$. The projected token sequence is then modeled by the BSS to obtain the behavioral feature $\mathbf{F}_{\mathrm{BSS}}\in\mathbb{R}^{B\times T\times d}$.

Finally, the LIS and BSS features are integrated by mean fusion to obtain the frame-level identity representation $\mathbf{F}=\frac{1}{2}\left(\mathbf{F}_{\mathrm{LIS}}+\mathbf{F}_{\mathrm{BSS}}\right)$. The video-level identity representation is obtained by temporal average pooling, i.e., $\mathbf{f}=\frac{1}{T}\sum_{t=1}^{T}\mathbf{F}_{t}$, where $\mathbf{f}\in\mathbb{R}^{B\times d}$. During training, the final identity feature is supervised by the AM-Softmax loss to enhance intra-class compactness and inter-class separability. During inference, the classification head is removed, and the extracted feature is used for similarity matching between enrollment and verification samples.

\subsection{Behavioral Sequential Stream}

The BSS contains two complementary temporal difference Mamba branches, namely the 1st-TD Mamba Branch and the 2nd-TD Mamba Branch. Given the projected token sequence $\mathbf{Z}_{\mathrm{proj}}=\{\mathbf{z}_1,\mathbf{z}_2,\ldots,\mathbf{z}_T\}$, where $\mathbf{z}_t\in\mathbb{R}^{B\times N\times D}$, the 1st-TD Mamba Branch constructs the first-order temporal difference sequence as $\Delta\mathbf{z}_t=\mathbf{z}_{t+1}-\mathbf{z}_t$. The first-order difference mainly describes velocity-like behavioral variations, such as finger movement speed, wrist rotation tendency, and local motion direction. Similarly, the 2nd-TD Mamba Branch constructs the second-order temporal difference sequence as $\Delta^2\mathbf{z}_t=\mathbf{z}_{t+2}-2\mathbf{z}_{t+1}+\mathbf{z}_t$. The second-order difference describes changes in motion variations, such as acceleration, deceleration, and motion turning patterns during gesture execution.

To model the behavioral sequence of each spatial location, each temporal difference sequence is reshaped along the spatial-token dimension and then normalized before being fed into stacked bidirectional Mamba layers. Specifically, for the $k$-th branch, where $k\in\{1,2\}$, the reshaped sequence has the size $\hat{\mathbf{S}}^{k}\in\mathbb{R}^{(BN)\times L_k\times D}$, where $L_1=T-1$ and $L_2=T-2$. Bidirectional Mamba captures both forward and backward temporal contexts, which helps model the complete gesture execution process. After another $LayerNorm$ layer and temporal average pooling, the spatial behavioral token set of the $k$-th branch is obtained as $\bar{\mathbf{H}}^{k}=[\bar{\mathbf{h}}_{1}^{k},\bar{\mathbf{h}}_{2}^{k},\ldots,\bar{\mathbf{h}}_{N}^{k}]$, where $\bar{\mathbf{h}}_{i}^{k}\in\mathbb{R}^{B\times D}$ denotes the behavioral feature of the $i$-th spatial token.

We further propose the Attention-Guided Behavioral Token Aggregation (ABTA) module to adaptively aggregate spatial behavioral tokens. For the $k$-th branch, ABTA maps $\bar{\mathbf{H}}^{k}\in\mathbb{R}^{B\times N\times D}$ into $\mathbf{f}_{\mathrm{BSS}}^{k}\in\mathbb{R}^{B\times D}$. It first computes a mean-pooled feature $\mathbf{f}_{\mathrm{mean}}^{k}=\frac{1}{N}\sum_{i=1}^{N}\bar{\mathbf{h}}_{i}^{k}$ to preserve global behavioral information. Meanwhile, a learnable positional embedding $\mathbf{E}_{\mathrm{pos}}\in\mathbb{R}^{1\times N\times D}$ is added for attention score prediction, and the attention weights are computed as $\boldsymbol{\alpha}^{k}={Softmax}\left(g\left(\bar{\mathbf{H}}^{k}+\mathbf{E}_{\mathrm{pos}}\right)\right)$, where $g(\cdot)$ consists of $LayerNorm$, $Linear$, and $GELU$ layers. The attention-pooled feature is obtained by $\mathbf{f}_{\mathrm{att}}^{k}=\sum_{i=1}^{N}\alpha_i^{k}\bar{\mathbf{h}}_{i}^{k}$. Finally, ABTA outputs $\mathbf{f}_{\mathrm{BSS}}^{k}=(1-\lambda)\mathbf{f}_{\mathrm{mean}}^{k}+\lambda\mathbf{f}_{\mathrm{att}}^{k}$, where $\lambda=\sigma(\theta)$ is a learnable mixing weight initialized as $0.1$. The two branch features are fused by mean fusion, i.e., $\mathbf{f}_{\mathrm{BSS}}=\frac{1}{2}\left(\mathbf{f}_{\mathrm{BSS}}^{1}+\mathbf{f}_{\mathrm{BSS}}^{2}\right)$.

\section{Experiments}
\subsection{Datasets and Protocols}
Experiments are conducted on the SCUT-DHGA benchmark dataset. The dataset contains 29,160 dynamic gesture videos from 193 subjects. The testing videos were collected in two sessions with an interval of about one week, forming two evaluation settings: Single and Cross. The Cross setting is more challenging as it reflects cross-session authentication variations.

SCUT-DHGA contains 6 dynamic gestures. This paper primarily conducts experiments under the MG and UMG settings of the Define Hand Gesture Authentication (DGA) protocol. As shown in Table \ref{tab:protocols}, MG uses all gestures for training and uses a specified gesture $G_i$ for registration and verification; UMG excludes $G_i$ during training, using only the remaining gestures, while registration and verification still use $G_i$. Therefore, UMG can evaluate a model's ability to generalize to unseen gestures. The evaluation metric is the equal error rate (EER), with a lower EER indicating better authentication performance.

\begin{table}[t]
	\centering
	\caption{Experimental protocols used in this work on the SCUT-DHGA dataset. ``$G_i$'' denotes the $i$-th hand gesture.}
	\label{tab:protocols}
	\small
	\begin{tabular}{c c c c}
		\hline
		\multirow{2}{*}{Experiments} 
		& \multirow{2}{*}{Training Gestures} 
		& \multicolumn{2}{c}{Testing Gesture} \\
		\cline{3-4}
		& & Enrollment & Verification \\
		\hline
		MG$i$  & All gestures & $G_i$ & $G_i$ \\
		UMG$i$ & All gestures except $G_i$ & $G_i$ & $G_i$ \\
		\hline
	\end{tabular}
\vspace{-0.5cm}
\end{table}

\subsection{Implementation Details}


ResNet18 is adopted as the shared visual encoder and initialized with ImageNet-pretrained weights. The Mamba hidden dimension, state dimension, and expansion ratio are set to 256, 16, and 2, respectively. Each video is uniformly sampled into $T=20$ frames, and each frame is resized to $224\times224$. The model is trained using Adam optimizer with a batch size of 6 and a dropout rate of 0.5. The learning rate is set to $1\times10^{-4}$ for the fully connected layer $\mathrm{FC}_{\mathrm{LIS}}$ and the learnable AM-Softmax weight matrix, and to $1\times10^{-5}$ for the remaining parameters. The former is reduced by 0.1 at the 25th epoch, and both are multiplied by 0.8 at the 50th epoch. Random rotation and color jittering are used for data augmentation. All models are trained for 50 epochs and then evaluated.

\begin{table*}[t]
	\centering
	\caption{Comparison of our approach with other networks.}
	\label{table:comparison_SOTA}
	\scriptsize
	\begin{tabular}{ccccccccccc}
		\toprule
		\multirow{3}{*}{Type} 
		& \multirow{3}{*}{Networks} 
		& \multicolumn{5}{c}{EER(\%)} 
		& \multirow{3}{*}{\begin{tabular}[c]{@{}c@{}}Parameters\\(M)\end{tabular}} 
		& \multirow{3}{*}{\begin{tabular}[c]{@{}c@{}}FLOPs \\(G)/Video \end{tabular}} \\ 
		\cmidrule(lr){3-7}
		& 
		& \multicolumn{2}{c}{UMG} 
		& \multicolumn{2}{c}{MG} 
		& \multirow{2}{*}{Average} 
		& 
		& \\ 
		\cmidrule(lr){3-4} \cmidrule(lr){5-6}
		& 
		& Single 
		& Cross 
		& Single 
		& Cross 
		& 
		& 
		& \\ 
		\midrule
		
		\multirow{6}{*}{\begin{tabular}[c]{@{}c@{}}3D\\CNN\end{tabular}} 
		& DHGA-net \cite{liu2020dynamic}     
		& 1.470 & 7.833  & 1.281 & 7.163  & 4.437 & 4.5  & 31.2  \\
		& R(2+1)D \cite{tran2018closer}      
		& 1.096 & 9.078  & 1.096 & 10.081 & 5.338 & 33.6 & 108.4 \\
		& ResNet3D-18 \cite{hara2018can}     
		& 0.611 & 7.326  & 0.459 & 6.696  & 3.773 & 33.5 & 117.6 \\
		& ResNet3D-101 \cite{wu2021spatiotemporal} 
		& 2.989 & 13.719 & 1.622 & 13.078 & 7.852 & 86.4 & 323.4 \\
		& SlowFast \cite{feichtenhofer2019slowfast} 
		& 3.374 & 9.807  & 3.870 & 13.107 & 7.540 & 34.9 & 34.9  \\
		& 3DTDS \cite{song2022video}         
		& 0.119 & 1.852  & 0.219 & 2.604  & 1.199 & 11.6 & 36.5  \\
		\midrule
		
		\multirow{11}{*}{\begin{tabular}[c]{@{}c@{}}2D\\CNN\end{tabular}} 
		& TDS \cite{song2021tds}             
		& 0.407 & 3.089  & 0.570 & 3.759  & 1.956 & 11.8 & 36.7  \\
		& TEA \cite{li2020tea}               
		& 0.689 & 8.400  & 0.519 & 7.567  & 4.294 & 9.0  & 28.9  \\
		& TEI \cite{liu2020teinet}           
		& 1.056 & 14.256 & 0.944 & 12.359 & 7.154 & 11.6 & 36.4  \\
		& TDN \cite{wang2021tdn}             
		& 0.807 & 4.015  & 0.500 & 3.522  & 2.211 & 11.8 & 41.6  \\
		& ResNet50 \cite{he2016deep}         
		& 0.359 & 3.378  & 0.615 & 3.644  & 1.999 & 24.6 & 82.2  \\
		& SE-ResNet \cite{hu2018squeeze}     
		& 0.800 & 4.807  & 0.552 & 3.448  & 2.402 & 27.2 & 82.4  \\
		& ResNet18 \cite{he2016deep}         
		& 0.096 & 1.796 & 0.144 & 2.363 & 1.100 & 11.5 & 36.4 \\
		& GaitSet \cite{chao2021gaitset}     
		& 0.152 & 2.530  & 0.178 & 2.078  & 1.235 & 22.9 & 38.2 \\
		& DeepGaitV2 \cite{fan2023deepgaitv2} 
		& 0.400 & 3.656  & 0.333 & 2.370  & 1.690 & 6.3  & 495.3 \\
		& TS-CNNs \cite{wu2016two}           
		& 2.270 & 9.678  & 2.948 & 10.741 & 6.409 & 122.5 & 28.6 \\
		& TS-DHGANet \cite{lin2025multiscale}                    
		& 0.133 & 2.559  & 0.059 & 1.456 & 1.052 & 7.8 & 27.2 \\
		\midrule
		
		\multirow{2}{*}{Mamba}
		& Mamba \cite{gu2023mamba}            
		& 11.159 & 24.889 & 11.893 & 29.552 & 19.373 & 93.3 & 92.6 \\
		& PhysMamba \cite{luo2024physmamba}                       
		& 1.196  & 10.485 & 0.941  & 12.615 & 6.309  & 0.8  & 24.6 \\
		\midrule
		
		2D CNN+Mamba 
		& MambaHand(Ours)                   
		& \ul{0.144} & 1.930  & 0.148 & \ul{1.715}  & \textbf{0.984} & 13.7 & 38.4 \\
		
		\bottomrule
	\end{tabular}
	\vspace{-0.3cm}
\end{table*}

\subsection{Comparison with Other Methods}

As shown in Table~\ref{table:comparison_SOTA}, the proposed MambaHand Net outperforms representative 3D and 2D-CNN-based methods. In particular, compared with the previous SOTA TS-DHGANet, MambaHand Net achieves competitive results under both UMG-Cross and MG-Cross settings. The results demonstrate the advantage of our method in modeling long-range behavioral patterns over 3D CNNs and explicit motion representations (e.g., optical flow and dynamic images).

Meanwhile, directly applying Mamba or Mamba-based models does not achieve satisfactory performance on the DHGA task. This suggests that DHGA can not simply rely on existing sequence models but requires task-specific designs for spatial token extraction, multi-order behavioral sequence construction, and identity-discriminative token aggregation. In contrast, MambaHand Net benefits from the collaborative design of the shared CNN encoder, LIS, BSS, and ABTA, achieving better performance while maintaining moderate parameters and computational cost.

\begin{table}[t]
	\centering
	\begin{minipage}[t]{0.65 \textwidth}
		\centering
		\caption{Ablation study of the proposed components.}
		\label{table:ablation}
		\small
		\begin{tabular}{cccc}
			\toprule
			\multirow{2}{*}{Method} 
			& \multirow{2}{*}{Network} 
			& \multicolumn{2}{c}{EER(\%)} \\
			\cmidrule(lr){3-4}
			& & Single & Cross \\
			\midrule
			\textit{Sche. 1} & LIS Only              & 0.144 & 2.363  \\
			\textit{Sche. 2} & BSS Only              & 1.078 & 12.900 \\
			\textit{Sche. 3} & LIS+1st-TD Branch   & 0.163 & 2.285  \\
			\textit{Sche. 4} & LIS+2nd-TD Branch   & 0.096 & 2.352  \\
			\textit{Sche. 5} & w/o $\mathbf{E}_{\mathrm{pos}}$ & 0.167 & 1.726  \\
			\textit{Sche. 6} & LIS+BSS (Ours)      & 0.148 & \textbf{1.715} \\
			\bottomrule
		\end{tabular}
	\end{minipage}
	\hfill
	\begin{minipage}[t]{0.3 \textwidth}
		\centering
		\caption{Effect of BSS connection stages.}
		\label{table:bss_connection_stage}
		\small
		\begin{tabular}{@{}ccc@{}}
			\toprule
			\multirow{2}{*}{Method} 
			& \multicolumn{2}{c}{EER(\%)} \\
			\cmidrule(lr){2-3}
			& Single & Cross \\
			\midrule
			After $L1$ & 0.185 & 2.163 \\
			After $L2$ & \textbf{0.148} & \textbf{1.715} \\
			After $L3$ & 0.163 & 2.519 \\
			After $L4$ & 0.204 & 1.974 \\
			\bottomrule
		\end{tabular}
	\end{minipage}
\end{table}

\subsection{Ablation Experiments}
\paragraph{\textbf{Effectiveness of the Proposed Components.}}
As shown in Table~\ref{table:ablation}, compared with using only LIS (\textit{Sche. 1}), only BSS (\textit{Sche. 2}), or a single TD branch (\textit{Sche. 3} and \textit{4}), the complete model achieves better overall performance, demonstrating the complementarity between local identity cues and sequential behavioral features. Removing the positional embedding (\textit{Sche. 5}) slightly degrades the performance, indicating that the learnable positional embedding provides useful spatial information for ABTA to aggregate identity-discriminative local dynamic regions. Overall, the complete model achieves the best Cross performance, verifying the effectiveness of the proposed components.


\paragraph{\textbf{Effect of BSS Connection Stage.}}

We connect the BSS after different ResNet18 stages, including $L1$, $L2$, $L3$, and $L4$. As shown in Table~\ref{table:bss_connection_stage}, using the feature after $L2$ achieves the best performance. This indicates that intermediate features are more suitable for behavioral sequence modeling. Shallow $L1$ features preserve more spatial details but lack sufficient semantic representation, while deeper $L3$ and $L4$ features contain stronger semantics but lose fine-grained local motion cues. In contrast, $L2$ features provide a better balance between spatial details and semantic information.


\begin{table}[t]
	\centering
	\begin{minipage}[t]{0.46\textwidth}
	\centering
	\caption{Effect of the number of Mamba blocks in each branch.}
	\label{table:mamba_block_number}
	\small
	\begin{tabular}{@{}ccc@{}}
		\toprule
		\multirow{2}{*}{\begin{tabular}{c}Mamba Block\\per Branch\end{tabular}}
		& \multicolumn{2}{c}{EER(\%)} \\
		\cmidrule(lr){2-3}
		& Single & Cross \\
		\midrule
		1 & 0.156 & 2.707 \\
		2 & 0.148 & \textbf{1.715} \\
		3 & 0.133 & 1.796 \\
		4 & 0.163 & 1.989 \\
		5 & 0.152 & 2.648 \\
		\bottomrule
	\end{tabular}
	\end{minipage}
	\hfill
	\begin{minipage}[t]{0.52\textwidth}
		\centering
	\caption{Effectiveness of Attention-Guided Behavioral Token Aggregation Strategy.}
	\label{table:token_aggregation_strategy}
	\small
	\begin{tabular}{ccc}
		\toprule
		\multirow{2}{*}{Method}
		& \multicolumn{2}{c}{EER(\%)} \\
		\cmidrule(lr){2-3}
		& Single & Cross \\
		\midrule
		Mean Pool      & 0.178 & 3.141 \\
		Sum Pool       & 0.185 & 2.652 \\
		Max Pool       & 0.122 & 2.378 \\
		Attention Pool & 0.170 & 2.007 \\
		ABTA           & 0.148 & \textbf{1.715} \\
		\bottomrule
	\end{tabular}
\end{minipage}
\vspace{-0.5cm}
\end{table}

\paragraph{\textbf{Effect of Number of Mamba Blocks.}}
As shown in Table~\ref{table:mamba_block_number}, increasing the number of Mamba blocks per BSS branch moderately improves long-range behavioral sequence modeling. However, further increasing the depth degrades performance, possibly because an overly deep temporal modeling structure introduces redundant parameters and tends to overfit training hand gestures, reducing cross-session generalization. Therefore, using two Mamba blocks in each branch provides the best balance between performance and model complexity.

\paragraph{\textbf{Effectiveness of ABTA.}}
We compare our ABTA with several token aggregation strategies. As shown in Table~\ref{table:token_aggregation_strategy}, ABTA achieves the best overall performance. By combining mean aggregation and attention-guided aggregation, ABTA preserves global behavioral information while emphasizing discriminative local dynamic regions, thereby improving the robustness of cross-session dynamic hand gesture authentication.

\section{Conclusion and Future Work}
This paper presented MambaHand Net, a two-stream CNN-Mamba framework for dynamic hand gesture authentication. The proposed framework combines CNN-based local identity modeling with Mamba-based long-range behavioral modeling, while the ABTA module further enhances discriminative behavioral representation learning. Experiments on the SCUT-DHGA benchmark demonstrate SOTA average authentication performance and strong cross-session robustness. Future work will investigate lightweight Mamba architectures, multimodal authentication, and real-time mobile deployment.

\vspace{-0.5cm}

\subsubsection*{Acknowledgment.}
This work was supported by the National Natural Science Foundation of China under Grant 62376100, Guangdong Basic and Applied Basic Research Foundation under Grant 2024A1515110009 and Grant 2026A1515010038, Youth S\&T Talent Support Programme of Guangdong Provincial Association for Science and Technology under Grant SKXRC2026169. Corresponding author: Junduan Huang.

\bibliographystyle{splncs}
\bibliography{references}

\end{document}